\documentclass[conference,12pt,onecolumn]{IEEEtran}

\IEEEoverridecommandlockouts
\usepackage{amsmath,amssymb,amsfonts}
\usepackage{mathpazo}
\usepackage[ruled,vlined,linesnumbered]{algorithm2e}

\usepackage{pdfpages}

\def\BibTeX{{\rm B\kern-.05em{\sc i\kern-.025em b}\kern-.08em
    T\kern-.1667em\lower.7ex\hbox{E}\kern-.125emX}}

\newcommand\approach{\xspace{ }RTRA\xspace}
\newcommand\datasetOne{\xspace{ }iFood251\xspace}

\usepackage{bm}
\newcommand{\bt}{\bm{\theta}}

\begin{document}

\title{\approach: \textbf{R}apid \textbf{T}raining of \textbf{R}egularization-based \textbf{A}pproaches in Continual Learning}

% \author{\IEEEauthorblockN{ Sahil Nokhwal }
% \IEEEauthorblockA{\textit{Computer Science Department} \\
% \textit{University of Memphis} \\
% Memphis, TN, USA \\
% nokhwal.official@gmail.com}
% \and
% \IEEEauthorblockN{ Nirman Kumar }
% \IEEEauthorblockA{\textit{Computer Science Department} \\
% \textit{University of Memphis}\\
% Memphis, TN, USA \\
% nkumar8@memphis.edu}}

\author{\IEEEauthorblockN{Sahil Nokhwal}
\IEEEauthorblockA{\textit{Dept. Computer Science} \\
\textit{University of Memphis}\\
Memphis, USA \\
nokhwal.official@gmail.com}
~\\
\and
\IEEEauthorblockN{Nirman Kumar}
\IEEEauthorblockA{\textit{Dept. Computer Science} \\
\textit{University of Memphis}\\
Memphis, USA \\
nkumar8@memphis.edu}
}

\maketitle

\begin{abstract}
%The performance of neural networks on previous tasks is often affected when they undergo several tasks sequentially without access to previous data. Catastrophic forgetting has been identified as a significant impediment to the advancement of continual learning (CL). In regularization-based CL, important parameters are penalized for updating to prevent catastrophic forgetting by defining an approximation loss function based on prior tasks. Our efforts are likewise related to regularization and are based on Elastic Weight Consolidation (EwC)\cite{kirkpatrick2017overcoming}. The time and effort required to retrain a neural network often make its use impractical. Our goal is to improve the training of regularization-based methods without sacrificing test-data performance. To ensure the quality of our work, we compare the proposed\approach approach against EwC using the \datasetOne dataset. As shown by our findings, \approach has a clear edge over the state-of-the-art approach.
Catastrophic forgetting(CF) is a significant challenge in continual learning (CL). In regularization-based approaches to mitigate CF, modifications to important training parameters are penalized in subsequent tasks using an appropriate loss function. We propose the\approach, a modification to the widely used Elastic Weight Consolidation (EWC) regularization scheme, using the Natural Gradient for loss function optimization. Our approach improves the training of regularization-based methods without sacrificing test-data performance. We compare the proposed\approach approach against EWC using the \datasetOne dataset. We show that\approach has a clear edge over the state-of-the-art approaches.
\end{abstract}


\section{Introduction}
%Overfitting a neural network on the input data streaming is frequent since their hyperparameters are typically set quite high. Additionally, optimization algorithms like stochastic gradient descent (SGD) \cite{robbins1951stochastic} or its derivatives are known to have an "implicit regularization" \cite{gunasekar2018characterizing, gunasekar2018implicit, azizan2018stochastic, neyshabur2014search} property that allows these kinds of networks to generalize effectively regardless of overfitting \cite{bartlett2020benign, bartlett2021deep, belkin2018overfitting, belkin2019reconciling, nakkiran2021deep}. The process of regularization is frequently accomplished in the research literature using techniques like early stopping and an effective method for regulating the parameters of the network is to incorporate weight decay. The primary difficulty of using such methods, however, lies in the fact that the convergent abilities are often not known, and consequently, they often are not accompanied by assurances of efficiency.

Regularization is a common technique in Machine learning to minimize overfitting and underfitting of models. This is particularly important for neural network models that are prone to overfitting since their hyperparameters are typically set high. In the area of Continual Learning (CL), the regularization technique is important as it helps to minimize overfitting of the model to a new task, which would thereby cause catastrophic forgetting for the classes in older tasks. The EWC \cite{kirkpatrick2017overcoming} is a well-known approach that is based on the idea of regularization and controlling the deviation of important parameters (from the old model), during the retraining of a new model. In this paper, we propose and evaluate the use of the natural gradient (NG) in the EWC setting. As expected, the NG allows for faster retraining process and thus improves the overall training time and performance of such systems. This is especially useful for CL because the model retraining is done several times, and saving time on retraining is therefore important. The use of the NG in the EWC setting also has another compelling reason: The NG has the potential to improve the convergence rate of any optimization algorithm that is based on gradient descent (used in training most deep learning models), but the reason it is not employed is because it is expensive to compute. It computation relies on the inverse of the Fisher Information Matrix, and during EWC this matrix (more precisely a diagonal approximation to it), is computed anyway. Thus, this can be exploited to reap the benefits of NG in this setting.

We also use a food dataset to evaluate our method. While the CIFAR10, CIFAR100, and ImageNet datasets have all seen extensive study in CL, food classification datasets have been less investigated. The challenge is difficult since there are so many distinct types of food, many of which seem identical, and there aren't enough huge datasets available for training deep models. Therefore, exploring the issue of food classification problems in continual machine learning holds great importance. Here are our specific contributions:
\begin{enumerate}
    \item We propose the use of the natural gradient (NG) in a regularization-based class-incremental learning (CIL) setup to train a neural network faster while retaining the model's accuracy. As far as we know, this is the first study to use the NG in a CL setting. 
    \item We propose new benchmarks for the\datasetOne dataset, that has not been researched yet in the class-incremental learning domain.
\end{enumerate}

\section{Related Work}
A plethora of continual learning approaches have been presented lately as a solution to the issue of catastrophic forgetting. Three main types of CL methods to mitigate catastrophic forgetting (CF) are as follows:

\subsection{Regularization-based continual learning approaches}
%Regularized learning belongs to the first class of techniques. 
In order to prevent CF in artificial neural networks, Elastic Weight Consolidation (EWC) \cite{kirkpatrick2017overcoming} demonstrates how synaptic consolidation may be tailored to a current task, allowing it to keep track of the relevant weights from prior tasks and selectively modify their plasticity. A comparable online importance score during a whole learning curve is computed by Synaptic Intelligence \cite{zenke2017continual}. Other modifications of EWC\cite{nokhwal2023rtra} have also been studied.

%A modification of Elastic Weight Consolidation (EWC), known as RotateEWC (REWC) \cite{liu2018rotate}, is a method that estimates the diagonal components of Fisher information matrix (FIM) of model parameters. This estimation is performed by computing the factorized rotation over the parameter space.

Other methods include choosing parameters specific to a particular task. Knowledge distillation is used in Learning without Forgetting (LwF) \cite{li2017learning} to impose similarity between the model and the current task's soft descriptors from earlier acquired tasks. \cite{jung2018less} involves regularizing the difference in $L_2$ among the last hidden layer activations of the task at hand and parameters of earlier trained tasks.

\subsection{Architecture-based continual learning approaches}
These techniques involve expanding the bandwidth of the network. The Progressive neural network (PGN) \cite{rusu2016progressive} widens the model structure by assigning separate models with constant memory size to train along incoming input, hence prohibiting updates to previously trained models on earlier-learned tasks. See also PathNet \cite{fernando2017pathnet} and DEN (dynamically expanding network) \cite{yoon2017lifelong}.
%To identify which sections of a model may be repurposed to acquire future tasks, PathNet \cite{fernando2017pathnet} isolates task-specific routes to prevent CF. In order to continuously learn novel tasks, the Dynamically Expanding Network (DEN) \cite{yoon2017lifelong} chooses synapses periodically in order to modify or extend synapse ability via the use of cluster sparse regularization.

\subsection{Rehearsal-based continual learning approaches}
To effectively remember prior task knowledge, these approaches make use of rehearsal memory, in which previously learned task exemplars are retained. A number of studies have been conducted on such models, including iCaRL, perhaps the most widely known of them, and a few related articles \cite{rebuffi2017icarl, nokhwal2023dss,nokhwal2023pbes, nokhwal2024survey}. %Learning a limited number of associations to a group of tasks regardless of utilizing task labels is what the partial derivative of episodic memory (GEM) \cite{lopez2017gradient} does. Additionally, GEM demonstrates the ability to achieve a positive transfer of learning to new tasks and mitigate the negative interference on previously acquired tasks. The iCaRL framework, as introduced by \cite{rebuffi2017icarl}, incorporates the utilization of the cross-entropy classification loss of the current task with distillation loss for previously acquired tasks. The method employed a KNN classifier along with selecting exemplars from every task based on the proximity of their embeddings to the mean data point of their respective classes\cite{nokhwal2023pbes}.

\section{Problem formulation}
\label{sec:prob_statement} 
%The configuration can possibly be expressed in the following manner. 
The incoming stream of data for a class-incremental learning (CIL) setup is denoted as $(x_1, y_1), \\ (x_2, y_2), \ldots, $ where $y_i$ represents the class label assigned to the data point $x_i$. The stream is theoretically partitioned into tasks. In the CIL scenario, the task identifier (task ID) is inherently absent throughout the inference process. In our methodology, the demarcation of task boundaries is determined by the process of aggregating data into batches. Each batch of data serves as a defining unit for a certain task, whereby the model is retrained. Although an option to implicitly track a task ID exists, this is not employed in the current CIL approach.

In a disjoint contextually class-incremental situation, the set of classes observed during distinct tasks is disjoint. Although the number of classes in every new task remains constant, there may be a disparity in the quantity of data points observed for each class. This is often referred to as \emph{data imbalance}. Hence, the overall quantity of data points may differ across different tasks.
%\begin{equation*} 
%   disjoint\_CIL \implies \bigcap ^ T_{t=0} = \Phi
%\end{equation*}
%The, $t$ represents an arbitrary task, whereas the total tasks are represented by $T$.

\section{General approach of  regularization-based CL to tackle catastrophic forgetting}
\begin{figure}[htbp!]
    \centering   
    \includegraphics[width=0.5\linewidth]{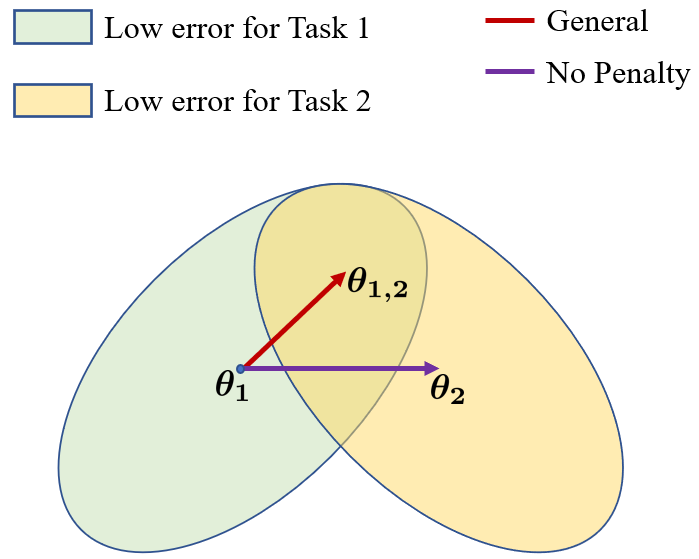}
    \caption{General approach of regularization-based CL}
    \label{fig:reg_fig}%
\end{figure}

Consider a model $M$ for image classification tasks that have been trained on a group of classes, from task $t_1$. Suppose we now need to update the model on another new task, task $t_2$, so that it can adequately perform on data points from classes in $t_2$ without significantly diminishing the original accuracy on data points from classes in $t_1$. A common approach is to retrain $M$ using the datasets from both tasks together; however, this is not always possible and, even if it is, that approach can be very computationally expensive depending on what task $t_1$ is. Particularly, in the conditions of continual learning, since the stream of training data for $t_1$ is not stored in its entirety, this problem is often given the restriction that the dataset for the task $t_1$ is not accessible.

However, the model parameters for $M$ after training on $t_1$ implicitly remember the task data, and thus in regularization-based approaches the goal is to minimally change parameters during retraining for $t_2$.

For example, in Figure~\ref{fig:reg_fig} the optimal parameters after training for $t_1$ are $\bt_1$ in the parameters space. While a (non-regularized) optimization for training on $t_2$ would move them to $\bt_2$, a regularized one moves to $\bt_{1,2}$ (for example) that enables good performance on classes from both tasks $t_1$ and $t_2$.

The way to achieve this is to penalize a change to the parameters. As such, a surrogate loss term is added to the existing cost function which will penalize the change in parameters for task $t_1$. This surrogate loss function is also usually weighted by the importance of various parameters, and modifications to more important ones are penalized more than modifications to the lesser important parameters. A typical equation after adding the surrogate loss term can be formulated as: 
\begin{equation*}
    \mathcal{\tilde{\mathcal{L}}}_{regularized}(\bt) = \mathcal{L}_{original}(\bt) + \varepsilon \sum_i \text{Penalty}(i),
\end{equation*}
where $\bt=(\theta_1, \ldots, \theta_i, \ldots)$ is a vector of parameters, $\mathcal{\tilde{\mathcal{L}}}_{regularized}(\bt)$ refers to the final loss after adding the current cross-entropy loss with the surrogate loss, and $\varepsilon$ denotes a constant to control the regularization effect while training the neural network. The cross-entropy (original) loss can be formulated as:
\begin{equation*}\label{eq:crossLoss}
    \mathcal{L}_{original}(\bt) = \sum_x \sum_{j = 1}^{r + s} -\hat{y}^{(j)} \log\left[p^{(j)}\right],
\end{equation*}
where $\hat{y} = (\hat{y}^{(1)}, \ldots, \hat{y}^{(r+s)})$ denotes a predicted one-hot encoding for a data point, $r$ represents the number of training classes that model is already trained on (until $(t-1)$ tasks), and $s$ is the total classes in the current task $t$. The predicted logits are $p = (p^{(1)}, p^{(2)}, \ldots, p^{(r+s)})$. (The $\hat{y}$ and $p$ of course depend on $\bt$.)

The $\text{Penalty}(i)$ denotes the penalty for changing the parameter $i$, and it is usually defined as,
\begin{equation}
\text{Penalty}(i) = \text{Importance}(i) \times (\text{Deviation of } \theta_i \text{ from } \theta_{1i}),
\label{eq:imp}%
\end{equation}
where $\text{Importance}(i)$ denotes the importance of parameter $\theta_i$. The computation of the importance and the deviation term above is discussed in the next section.

%NK - what is $L_2$ term - undefined
%One method is to use the $L_2$ term as a surrogate loss and is used in conjunction with the original loss term, which penalizes all parameters in the neural network equally. The assumption here is that all parameters matter equally, which is generally not true for neural networks. 

%Therefore, an intuitive method should be used to penalize modifying the important parameters of older tasks so that the changes in those parameters are minimal, while the changes in less important parameters could be substantial. 
%Therefore, when another task is trained, crucial parameters from older tasks are either not updated or updated minimally. Now, the question arises, how do we determine which parameters are most important and which are least important? The utilization of the Fisher Information Matrix (FIM) methodology is widely prevalent in academic literature for the purpose of quantifying the significance of parameters within a neural network.
%The posterior distribution of the parameters is used in a well-known research paper in this field to estimate parameter importance based on Bayesian analysis  \cite{kirkpatrick2017overcoming}.

\subsection{The Fisher Information Matrix and calculation of importance score}
The utilization of the Fisher Information Matrix (FIM) methodology is widely prevalent in academic literature for the purpose of quantifying the significance of parameters within a statistical model (such as a neural network). Using FIM \cite{geisser1992introduction}, SI \cite{li2017learning} and EWC \cite{kirkpatrick2017overcoming} finds key parameters in a model.

From the FIM, a parameter $\theta_i$'s importance score can be calculated as $\text{Importance}(i) = I_{ii}$,
%\begin{equation*}
    %Importance(i) = I_{\theta_i} \times \theta_i^2
    %\text{Importance}(i) = I_{ii}, % \times (\theta_i - \theta_{1i})^2
%\end{equation*}
where $I = [I_{ij}]_{n \times n}$ is the FIM (see below), and $n$ is the parameter count. The deviation term in Eq.~\ref{eq:imp} is usually the $i$th contribution to the importance weighted squared $\ell_2$ distance between $\bt$ and $\bt_1$, i.e., $(\theta_i - \theta_{1i})^2$. Thus, $\text{Penalty}(i) = I_{ii} (\theta_i - \theta_{1i})^2$.
The aforementioned metric integrates the sensitivity of the loss function to changes in the parameter, as quantified by the FIM, along with the absolute value of the parameter.
%\subsection{Calculation of Fisher Information(FI):}
%The Fisher Information is a numerical indicator that serves for the assessment of the quantity of data pertaining to a model parameter vector $\bt = (\theta_1, \ldots, \theta_n)$, for instance, the true mean, of a random variable within the presumed probability distribution. 
The FIM %for an individual parameter $\theta_i$, of a neural network model with regard to the data distribution can be computed as: 
for a parameter vector, $\bt$ of a neural network model with regard to the data distribution can be computed as: 
\[
I(\bt) = \begin{bmatrix}
I_{11} & I_{12} & \cdots & I_{1n} \\
I_{21} & I_{22} & \cdots & I_{2n} \\
\vdots & \vdots & \ddots & \vdots \\
I_{n1} & I_{n2} & \cdots & I_{nn}\\
\end{bmatrix},
 \]
where
\[ 
    I_{ij} = \mathbb{E}\left[ \frac{\partial \ln f(X; \bt)}{\partial \theta_i} \frac{\partial \ln f(X; \bt)}{\partial \theta_j} \right],
\]
and \( f(X; \bt) \) is the (unknown) probability density function (pdf) of the observable random variable \( X \). The diagonal elements of an FIM, denoted as $I_{ii}$, quantify the extent to which each parameter $\theta_i$ accounts for the entropy of $I_{\bt}$. Usually, the FIM is computed from samples to estimate the expectation, see \cite{martens_ng}.

The equation presented quantifies the degree of sensitivity exhibited by the logarithmic likelihood of the actual label $y$ in response to variations in the value of the parameter $\theta_i$. In the event of high sensitivity, a minor alteration in $\theta_i$ would result in a significant alteration in the logarithmic likelihood. This observation suggests that $\theta_i$ plays a crucial role in enabling the model to generate precise predictions.

\subsection{Diagonal approximation of Fisher Information Matrix}
The computation of a full FIM is considered infeasible, particularly in scenarios where there is a large number of parameters (often reaching millions). Hence, in the literature, there exists a plethora of methods to estimate an FIM. The most popular ones are using a diagonal approximation and diagonal-band approximation. %However, one direct approach involves approximating the FIM by utilizing its diagonal elements.
The process of approximating an FIM by considering only its diagonal is highly efficient. In particular, in its use for NG (see below), we need to compute the inverse and this is efficient for a diagonal approximation.

\section{Proposed technique}

\begin{figure*}[ht!]
    \centering    
    \includegraphics[width=0.95\linewidth]{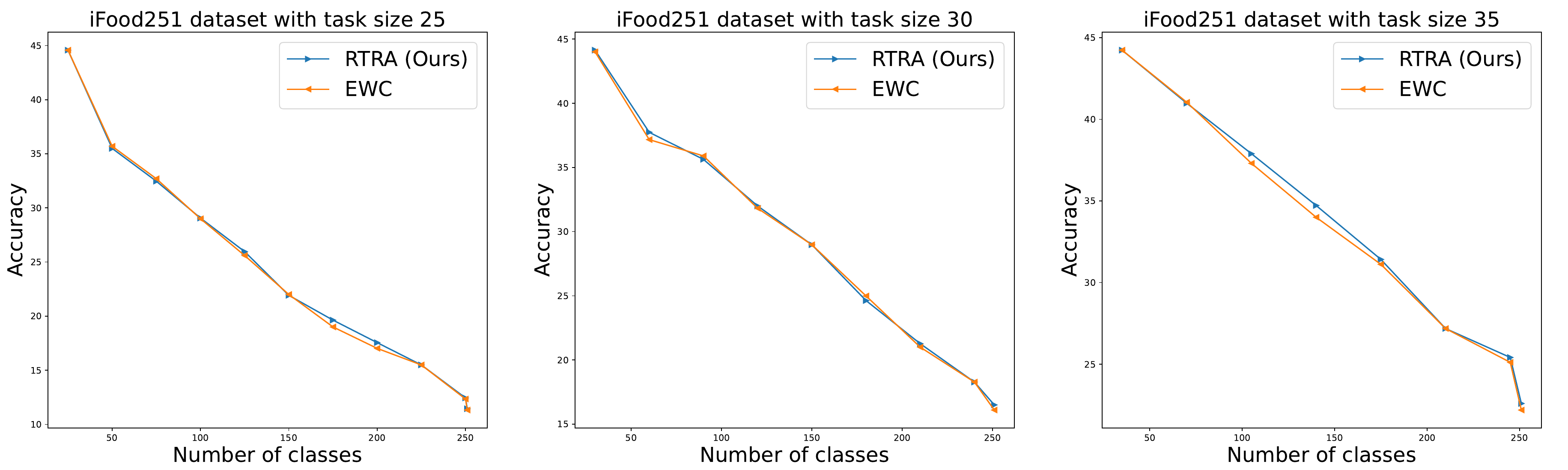}
    \caption{Per task accuracy obtained on\datasetOne dataset}
    \label{fig:results1}
\end{figure*}

% \begin{figure*}[ht!]
%     \centering    
%     \includegraphics[height=5cm, width=0.9\linewidth]{figures/results2.png}
%     \caption{Per task accuracy obtained on \datasetOne dataset}
%     \label{fig:results2}
% \end{figure*}

\begin{figure*}[ht!]
    \centering
    \includegraphics[width=0.6\linewidth]{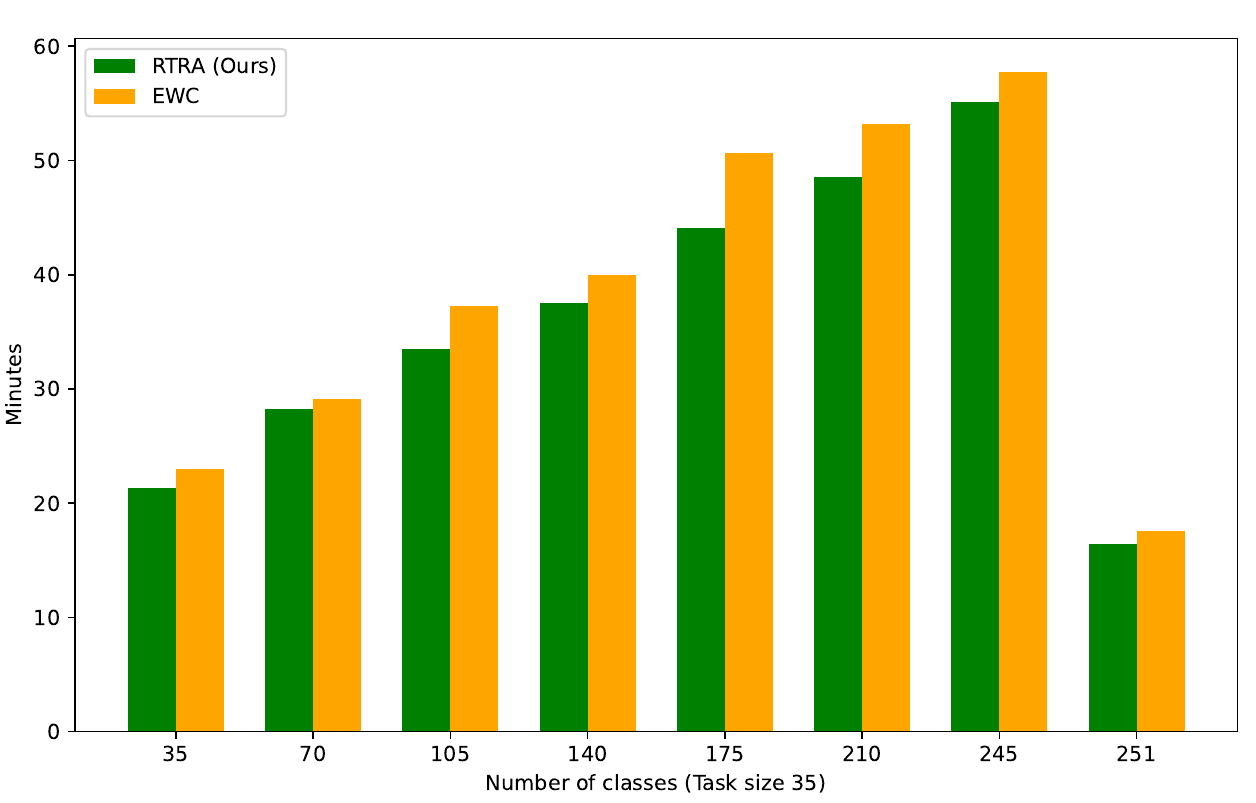}
    \caption{Comparison of time required between\approach (Ours) and the EWC techniques for task size 35}
    \label{fig:results_time}
\end{figure*}

We now discuss our proposed modification to the optimization method used to retrain the neural network that is based on Natural Gradients.
\begin{algorithm}
    \caption{Natural Gradient Descent Optimizer}\label{algo:ngd}
    \DontPrintSemicolon
    \SetKwFunction{FTrainer}{NGOptimizer}
    \SetKwProg{Fn}{Function}{:}{}
    \Fn{\FTrainer{$\bt^{t}, \eta$}}{
        \KwIn{model: $\bt^{t}$, $\eta$ is the learning rate}
        \BlankLine
        $I \leftarrow$ compute the FIM using \cite{kirkpatrick2017overcoming} \;
        %$I^{-1}$ $\leftarrow$ calculate the inverse of $I$ \;             
        $\bt^{t} \gets \bt^{t} - \eta I^{-1}  \nabla \mathcal{L}(\bt^{t}) $ \;
    }
\end{algorithm}

\subsection{Natural gradient descent (NGD)}
The Natural Gradient (NG) based algorithm only changes the definition of the gradient. It can be used in conjunction with any optimization algorithm based on the idea of gradient descent in general. The intuition is that the natural gradient uses a more \emph{natural} distance notion between the two \emph{distributions} given by parameter vectors $\bt$ and $(\bt + d \bt)$. While the $\ell_2$ norm is a proper metric, it depends on the parametrization ($\bt$) of the distribution and there could be multiple such parametrizations. On the other hand, a more natural measure would be some sort of distance between the distributions induced by the $\bt$ itself, as opposed to one on the parametrization. Such a notion is the KL divergence. Even though the KL divergence is not a metric on the space of distributions, it can be used to define the gradient, see \cite{martens_ng}. 

Another view of the NG is that one can view the space of distributions as a Riemannian manifold whose metric tensor is given by the FIM \cite{2018AcceleratingNG}. In this view, the entries of the FIM are viewed as the components of a Riemannian metric tensor that defines the quadratic form measuring the distance between two infinitesimally close points $\bt$ and $(\bt + d\bt)$, using the KL divergence (that can serve as a distance locally). %Here the distance is defined using a more natural metric, the KL divergence, as opposed to the usual $\ell_2$ metric. The NG involves utilizing the geometric properties of the space of the parameter endowed with the Riemannian metric as determined using the FIM, in order to modify the gradient changes. 
The conventional approach of gradient descent involves taking iterative steps in the direction that corresponds to the most significant reduction in the loss function. Nevertheless, these procedures may exhibit inefficiencies if they fail to consider the curvature or geometry that makes up the underpinning space of the parameter. The NG process involves modifying the model’s parameters in such a manner that remains unaffected by the selection of coordinate systems used to describe the model \cite{Jahani2021FastAS}. The concept is exactly the same as using the intrinsic distance in Riemannian geometry, as opposed to the Euclidean distance that depends on the coordinate system. According to the reference \cite{Jahani2021FastAS}, employing the use of the NG has the potential to enhance the convergence rate of algorithms for optimization and bolster their stability.

\subsection{Updates using Natural Gradient descent}
The NG $\tilde{\mathcal{L}}(\bt)$ can be expressed as the inverse of FIM times the standard gradient of the function's loss with regard to the parameters \cite{amari1998natural}.

\begin{equation*}
    \tilde{\mathcal{L}}(\bt) = I(\bt)^{-1} \nabla \mathcal{L}(\bt).
\end{equation*}

Here, $\mathcal{L}(\bt)$  is the cost function that the model needs to minimize, $\nabla \mathcal{L}(\bt)$ is the standard gradient of $\mathcal{L}$ with respect to the parameters $\bt$, and $I(\bt)$ is the FIM. Therefore, the updated equation utilizing the NG becomes:

\[
\bt_{\text{new}} = \bt_{\text{old}} - \eta \tilde{\mathcal{L}}(\bt),
\]
where \( \eta \) is the learning rate.

\section{Experimental Results}
\subsection{Setup}
\emph{Implementation details:} The implementation of ResNet32 given in the original work is used. The learning rate is set to 0.001 and epochs to 300.

\emph{Dataset:} The iFood251 dataset \cite{kaur2019foodx} is used for our study. In 2019, the dataset was initially utilized to conduct a competition at the Computer Vision and Pattern Recognition (CVPR) conference. The 251 classes include a comprehensive range of meticulously categorized and curated food items, consisting of a total of 120,216 training pictures that have been systematically gathered from various online sources. A validation set of 12,170 pictures was used as test data because the labels for the test data were not supplied by the organizers of the competition.

% \subsection{Backbone} The implementation of ResNet32 given in the original work \cite{he2016deep} was used in this work. The ResNet32 architecture is constructed using three hyper-blocks, resulting in a total parameter count of 0.5 × 106. The Conv-4 backbone is constructed with three blocks, each consisting of convolution operations with padding equal to  1, a kernel size equal to 3, and a stride equal to 1. The convolution processes are then followed by Batch Normalization and ReLU activation function. Additionally, each block includes average pooling with a kernel size equal to 2 and a stride equal to 2. In the following block, consisting of 64 maps of features, the identical techniques were conducted, however with the inclusion of dynamic average pooling to compress the maps to a unit structure in the dimension of space. The conventional weight initialization method was used, and the weights of BatchNorm were set to 1 while the bias was set to 0. The dimensions of the feature representations of Conv-4 and ResNet32 are both 64.

\subsection{Metrics used}
The findings have been documented using per-task accuracy, denoted as $a_i$, which denotes the accuracy attained after training each individual task \cite{tanwer2020system, nokhwal2023embau,nokhwal2024hffa, nokhwal2024iiot, nokhwal2024gsp, nokhwal2024opt}. The performance has been quantified in terms of minutes. The per-task-accuracy can be written as: $\text{Per-task-accuracy} = a_{i}.$

\subsection{Results}
This work concentrates on improving the training speed of a model using Natural Gradient \cite{yang1997natural}, therefore training using NG and SGD (EWC) has been compared and results have been illustrated in graphs \ref{fig:results1} and \ref{fig:results_time}. We demonstrate that our methodology surpasses EWC, establishing it as a favorable option for expediting the training of a model. While our proposed algorithm has been shown to be effective against EWC, it can also be used with any other contemporary CL approach. Using\approach, training took 7.71\% less time as compared to EWC, without compromising accuracy.

The observed upward trend in time shown in Graph 2 can be attributed to the need for the model to assess its performance against both current and previously encountered test data for each subsequent task throughout the retraining process.  The abrupt reduction in the duration of the last task is attributed to the fewer classes involved, i.e., 6 as opposed to the 35 classes typically included in prior tasks.

\section{Conclusion}
In this study, we suggest the use of Natural Gradient in the regularization-based CIL framework as a means to enhance the efficiency of neural network training, while maintaining the integrity of testing accuracy. Our proposed methodology has the potential to enhance the efficiency of the training process, resulting in the ability to achieve the same level of accuracy in 7.71\% less time.

\bibliographystyle{IEEEtran}
\bibliography{bibliography}
\end{document}